\pdfoutput=1

\documentclass[11pt]{article}
\PassOptionsToPackage{table}{xcolor}
\usepackage{multirow}
\usepackage{needspace}

\usepackage[preprint]{acl}
\usepackage{amsmath}
\usepackage{amssymb}
\usepackage{times}
\usepackage{latexsym}
\usepackage{pifont}  
\definecolor{myblue}{RGB}{36, 92, 181}
\definecolor{myorange}{RGB}{230, 126, 34}
\definecolor{mypurple}{RGB}{155, 89, 182}
\definecolor{myred}{RGB}{231, 76, 60}
\definecolor{myteal}{RGB}{26, 188, 156}
\usepackage{algorithm}
\usepackage{algpseudocode} 
\usepackage{graphicx}
\usepackage{subfigure}    

\usepackage[T1]{fontenc}

\usepackage[utf8]{inputenc}
\DeclareUnicodeCharacter{2009}{\,}

\usepackage{microtype}

\usepackage{inconsolata}

\usepackage{graphicx}
\usepackage{environ} 
\usepackage{listings}
\usepackage{enumitem}
\usepackage{natbib}
\usepackage{hyperref}
\usepackage{tcolorbox}
\usepackage[table]{xcolor}    
\usepackage{booktabs}
\usepackage{fontawesome5} 
\usepackage{fancybox}   

\usepackage{authblk}
\tcbset{
  myfancybox/.style={
    enhanced,
    colframe=black!60!blue,
    colback=blue!3,
    boxrule=0.5pt,
    arc=2mm,
    outer arc=2mm,
    sharp corners=downhill,
    fonttitle=\bfseries\small,
    coltitle=black,
    attach boxed title to top left={yshift=-2mm, xshift=2mm},
    boxed title style={
      colframe=black!80!blue,
      colback=black!5,
      boxrule=0.5pt,
      sharp corners,
    },
    width=\linewidth,
    before skip=10pt,
    after skip=10pt,
    left=3mm,
    right=3mm,
    top=2mm,
    bottom=2mm,
    drop shadow south east with={shadow xshift=0.3ex, shadow yshift=-0.3ex}
  }
}

\title{Rethinking All Evidence: Enhancing Trustworthy Retrieval-Augmented Generation via Conflict-Driven Summarization}

\author{
  \textbf{Juan Chen}$^{1,2}$ \quad
  \textbf{Baolong Bi}$^{2}$\thanks{Corresponding author.}\thanks{Authors from affiliation$^{2}$ are also affiliated with: Key Laboratory of Network Data Science and Technology, ICT, CAS; State Key Laboratory of AI Safety; University of Chinese
Academy of Sciences.} \quad
  \textbf{Wei Zhang$^{3}$} \quad
  \textbf{Jingyan Sui$^{1,2}$} \\
  \textbf{Xiaofei Zhu$^{4}$} \quad
  \textbf{Yuanzhuo Wang$^{1,2}$} \quad
  \textbf{Lingrui Mei$^{2 {\dagger}}$}\quad
  \textbf{Shenghua Liu$^{2 {\dagger}}$}  \\
  $^1$University of Chinese Academy of Sciences \\
  $^2$Chinese Academy of Sciences \\
  $^3$National University of Defense Technology \\
  $^4$Chongqing University of Technology
}

\begin{document}
\maketitle
\begin{abstract}
Retrieval-Augmented Generation (RAG) enhances large language models (LLMs) by integrating their parametric knowledge with external retrieved content. 
However, knowledge conflicts caused by internal inconsistencies or noisy retrieved content can severely undermine the generation reliability of RAG systems.
In this work, we argue that LLMs should rethink all evidence, including both retrieved content and internal knowledge, before generating responses.
We propose \textbf{CARE-RAG} (\textbf{C}onflict-\textbf{A}ware and \textbf{R}eliable \textbf{E}vidence for RAG), a novel framework that improves trustworthiness through \textit{Conflict-Driven Summarization} of all available evidence.
CARE-RAG first derives parameter-aware evidence by comparing parameter records to identify diverse internal perspectives. It then refines retrieved evidences to produce context-aware evidence, removing irrelevant or misleading content. To detect and summarize conflicts, we distill a 3B LLaMA3.2 model to perform conflict-driven summarization, enabling reliable synthesis across multiple sources.
To further ensure evaluation integrity, we introduce a QA Repair step to correct outdated or ambiguous benchmark answers.
Experiments on revised QA datasets with retrieval data show that CARE-RAG consistently outperforms strong RAG baselines, especially in scenarios with noisy or conflicting evidence.
\end{abstract}

\begin{figure}[ht]
  \includegraphics[width=\columnwidth]{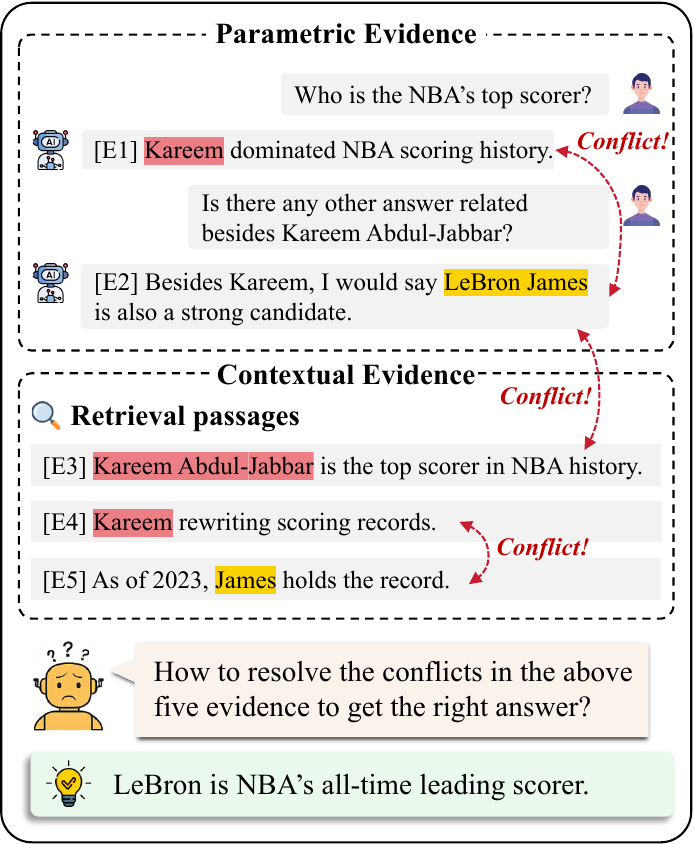}
  \caption{LLMs struggle to assess the reliability of evidence from different sources and to resolve conflicts among them, challenging the trustworthiness of RAG.}
  \label{fig:background_example}
  \label{fig:experiments}
\end{figure}
\section{Introduction}

Retrieval-Augmented Generation (RAG) has emerged as a powerful framework to equip large language models (LLMs)~\citep{achiam2023gpt, grattafiori2024llama} with access to external knowledge, enabling strong performance on knowledge-intensive tasks like question answering~\citep{karpukhin2020dense, guu2020retrieval, gao2023retrieval}. While RAG effectively extends the knowledge capacity of LLMs, its reliability in real-world applications remains a significant concern~\citep{santhanam2021colbertv2, fan2024survey}. 

RAG enhances LLMs' generation by leveraging both internal and external knowledge, but as shown in Figure \ref{fig:background_example}, it also introduces unreliable sources that make reasoning more difficult.
First, due to internal hallucinations~\citep{huang2023survey, tonmoy2024comprehensive}, LLMs often generate multiple inconsistent viewpoints for a given question. 
While introducing new retrieval contexts aims to supplement additional knowledge and alleviate these hallucinations, many retrieved evidences contain errors, noise, and even contradictions~\citep{yoran2023making, wang2023survey}. 
Moreover, the potential conflicts~\citep{xu2024knowledge, xie2023adaptive, shi2025ircan} between the model's internal parameter knowledge and the retrieved context further challenge the RAG generation process, where multiple knowledge sources interact in a black-box manner~\citep{bi2024factuality, mao2024fit}.

To address these issues, we propose that LLMs should rethink all evidence before generating responses in RAG framework, to clarify the relationships between the internal knowledge and the retrieved context. 
In this work, we introduce \textbf{CARE-RAG} (\textbf{C}onflict-\textbf{A}ware and \textbf{R}eliable \textbf{E}vidence for RAG), a novel framework that enhances the trustworthiness of RAG by synthesizing all available evidence based on conflict identification.

CARE-RAG first captures all evidence related to the query, sourced from both the LLM’s internal parameters and the retrieved documents. For the LLM’s internal knowledge, we generate parameter-aware evidence by comparing parameter records. Specifically, we concatenate the model’s previous generated parameter views and prompt the model to generate new perspectives, different from the existing ones, thereby covering all possible viewpoints to reduce internal hallucinations. For the retrieved documents, we perform fine-grained refinement to generate context-aware evidence, identifying and removing irrelevant noise. This reduces the risk of hallucinations caused by unrelated content, while also saving token usage, allowing the model to consider more context within token window limits and enhancing robustness.

While CARE-RAG explicitly lists all available evidence to ensure that as much relevant information as possible is considered, this also introduces more potential conflicts. To address this, we design a knowledge summarization step based on conflict detection, providing a final conflict report alongside all the evidence to guide the LLM. Specifically, we distill the capabilities of DeepSeek-v3 into a smaller LLaMA 3.2-3B model, enabling it to assess the conflict between two evidences and provide related reasoning. The distilled model efficiently cross-checks all evidence (both parameter-aware and context-aware) to detect conflicts and synthesize diverse knowledge perspectives. This additional information helps the LLM generate a reliable response based on all the input evidence.

We conduct experiments on five QA benchmarks—Natural Questions, TriviaQA, HotpotQA, ASQA, and WikiQA—covering both open-domain and multi-hop question answering. To improve supervision quality and ensure fairer evaluation, we introduce a lightweight answer-set augmentation procedure that corrects outdated or semantically inconsistent gold answers. This QA repair step is applied once before training and used consistently across all experiments. Results show that this augmentation leads to substantial gains in both EM and F1 across datasets. Compared to standard RAG, CARE-RAG with augmentation improves EM scores by up to 23.6\% (e.g., from 40.3 to 63.9 on NQ with LLaMA-3.2-8B), and outperforms the strongest existing baseline by an average of 3.8\% on EM. Further experiments confirm CARE-RAG’s robustness to the number of evidence and validate the effectiveness of each pipeline component, highlighting the importance of rethinking evidence in enhancing the RAG process.

Our main contributions are as follows:
\begin{itemize}
    \item We propose CARE-RAG, a novel framework for enhancing the trustworthiness of RAG by rethinking all available evidence via conflict-driven summarization.
    \item We perform QA repair on multiple widely-used QA datasets to ensure more accurate and reliable evaluation for the community. In addition, we distill and release a conflict detection model based on LLaMA-3.2–3B, capable of analyzing and identifying potential conflicts among input evidence.
    \item Experimental results show that CARE-RAG significantly improves the ability of LLMs to effectively integrate all available evidence, achieving state-of-the-art performance on multiple RAG tasks and demonstrating the importance of rethinking evidence in RAG process.
\end{itemize}

\begin{figure*}[!t]
  \includegraphics[width=\linewidth]{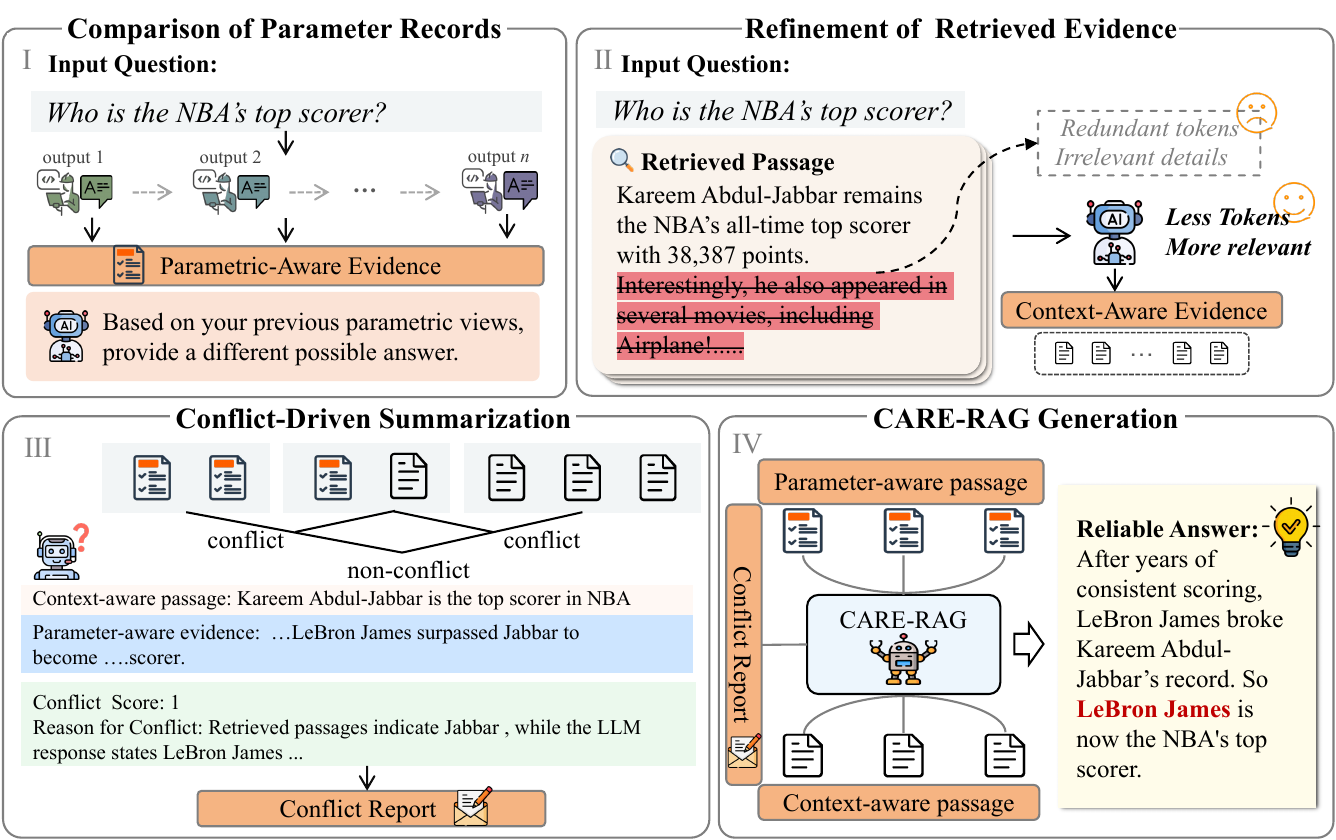}
  \caption{An illustration of CARE-RAG rethinking all available evidence via conflict-driven summarization. The framework consists of four stages:
    \textbf{(I) Comparison of parameter Records} licits and aggregates the model’s internal diverse perspectives into parameter-aware evidence;  
    \textbf{(II) Refinement of Retrieved Evidence} removes irrelevant noise from raw retrieved content to produce concise, context-aware evidence;  
    \textbf{(III) Conflict-Driven Summarization} detects and analyzes conflicts between parameter-aware and context-aware evidence; 
    \textbf{(IV) CARE-RAG Generation} synthesizes a final answer by reconciling conflicts and integrating all information.}
  \label{fig:framework}
\end{figure*}
\begin{algorithm}[!t]
\caption{CARE-RAG Inference Procedure}
\label{alg:srrag_detailed}
\begin{algorithmic}[1]
\Require Query \(q\); Retriever \(\mathcal{R}\); LLM \(\mathcal{M}\); Conflict detector \(\mathcal{M}_{c}\)
\Ensure Final answer \(\hat{a}\)
  \State \(\mathcal{E}_{p}\gets[]\)
  \State \(\Pi_{\dots}\) defined as above
  \vspace{0.5em}
  \State \textbf{parameter Record Comparison}
  \State \(a_{0}\gets\mathcal{M}(q;\,\Pi_{\text{init}})\)  
  \State \(\mathcal{E}_{p}.\mathrm{append}(a_{0})\)
  \For{\(i=1\) to \(n-1\)} 
    \State \(a_{i}\gets\mathcal{M}(q,\mathcal{E}_{p};\Pi_{\text{iter}})\)
    \State \(\mathcal{E}_{p}.\mathrm{append}(a_{i})\)
  \EndFor
  \State \(\mathcal{E}_{p}\gets\mathrm{merge}(\mathcal{E}_{p})\)
  \vspace{0.5em}
  \State \textbf{Retrieval Result Refinement}
  \State \(C\gets\mathcal{R}(q)\)
  \State \(\mathcal{E}_{c}\gets\mathcal{M}(q,C;\Pi_{\text{ref}})\)
  \vspace{0.5em}
  \State \textbf{Conflict-Driven Summarization}
  \State \((\delta_{c},r_{c})\gets\mathcal{M}_{c}(q,\mathcal{E}_{p},\mathcal{E}_{c};\Pi_{c})\)
  \If{\(\delta_{c}=1\)} 
    \State \(\mathcal{E}_{c}\gets\mathrm{augment}(\mathcal{E}_{c},r_{c})\)
  \EndIf
  \vspace{0.5em}
  \State \textbf{CARE-RAG Generation}
  \State \(\hat{a}\gets\mathcal{M}(q,\mathcal{E}_{p},\mathcal{E}_{c},\delta_{c},r_{c};\Pi_{\text{synth}})\)
  \State \Return \(\hat{a}\)
\end{algorithmic}
\end{algorithm}

\section{CARE-RAG: Conflict-Aware and 
Reliable Evidence for RAG}
\label{sec:methodology}
In this work, we propose CARE-RAG, a novel framework designed to enhance the trustworthiness of RAG systems. Unlike standard RAG that directly synthesize answers according to retrieved evidence in black-box manner, CARE-RAG introduces a four-stage framework that enables LLMs to thoroughly rethink all available evidence—both from parameter memory and retrieved context to generation. 
As illustrated in Figure~\ref{fig:framework}, CARE-RAG first derives parameter-aware evidence by comparing parameter records, thereby eliciting diverse internal perspectives.
It then refines the retrieved evidence to obtain context-aware evidence by removing irrelevant or noisy content. Finally, a distilled language model performs conflict-driven summarization to generate reliable answers by aggregating across multiple sources.
This framework explicitly separates the model’s parameter knowledge from external context, and mitigates hallucinations by resolving the complex conflicts between them. The detailed inference procedure of our CARE-RAG is presented in Algorithm~\ref{alg:srrag_detailed}.
\vspace{-5pt}
\subsection{Parameter Record Comparison}
\label{ssec:parameter_record_comparison}
Given a query \(q\), we first elicit the model’s parameter-aware evidence \(\mathcal{E}_p\) without retrieved context, aiming to establish its internal knowledge baseline before external evidence is introduced (as shown in Figure~\ref{fig:framework} Stage I). This involves:
\begin{subequations}
\begin{align}
  a_0 &\leftarrow \mathcal{M}(q;\,\Pi_{\text{init}})\,, \label{eq:init}\\
  a_i &\leftarrow \mathcal{M}\bigl(q,\mathcal{E}_p;\,\Pi_{\text{iter}}\bigr), 
    \quad i=1,\dots,n-1\,. \label{eq:iter}
\end{align}
\end{subequations}
Here, iterative prompting (Eq.~\ref{eq:iter}) systematically encourages the model to generate diverse internal perspectives, explicitly aiming to reduce internal hallucinations by capturing variability within its parameter knowledge. We then define
\[
  \mathcal{E}_{p} = \{a_{0},a_{1},\dots,a_{n-1}\},
\]
which encapsulates the model’s parameter-aware evidences, revealing potential internal inconsistencies or uncertainties.

\subsection{Retrieval Result Refinement}
\label{ssec:retrieval_result_refinement}
Concurrently, a retriever \(\mathcal{R}\) returns evidences \(C=\{c_{1},\dots,c_{k}\}\). To distill these into a concise \emph{context-aware evidence} \(\mathcal{E}_{c}\) focusing on salient information (as illustrated in Figure~\ref{fig:framework} Stage II), we use:
\begin{equation}
  \mathcal{E}_{c} \;\leftarrow\; \mathcal{M}(q,\,C;\,\Pi_{\text{ref}}),
\end{equation}
where \(\Pi_{\text{ref}}\) explicitly instructs the model to extract critical factual claims and eliminate irrelevant or redundant content. This refinement enhances the clarity and relevance of external evidence, facilitating subsequent conflict detection.
In addition, the refinement also saves token usage, allowing the model to consider more context within the token window and enhancing robustness.

\subsection{Conflict-Driven Summarization}
\label{ssec:conflict_driven_induction}
Given the parameter-aware evidences \(\mathcal{E}_{p}\) (internal knowledge) and the refined evidence \(\mathcal{E}_{c}\) (external knowledge), we explicitly identify discrepancies via a dedicated conflict detection module \(\mathcal{M}_{c}\) (Figure~\ref{fig:framework} Stage III):
\begin{equation}
  (\delta_{c},\,r_{c}) \;\leftarrow\;
  \mathcal{M}_{c}\bigl(q,\,\mathcal{E}_{p},\,\mathcal{E}_{c};\,\Pi_{c}\bigr),\quad
  \delta_{c}\in\{0,1\},
\end{equation}
where \(\delta_{c}=1\) indicates a conflict and \(r_{c}\) provides the natural-language rationale, forming a detailed "conflict report". Specifically, we construct a training dataset by annotating conflicts and their rationales using a teacher LLM (e.g., DeepSeek). We then distill this knowledge into a smaller, efficient LLaMA-3.2B model through supervised fine-tuning, enabling rapid and accurate conflict detection during inference.

\paragraph{No Conflict (\(\delta_{c} = 0\)).}
When no conflict is detected, the model primarily grounds its response in the refined external evidence \(\mathcal{E}_{c}\), while using the internal knowledge \(\mathcal{E}_{p}\) to provide additional support and increase confidence in the answer.

\paragraph{Conflict Detected (\(\delta_{c} = 1\)).}
When a conflict is identified, the model explicitly considers the rationale \(r_{c}\), critically evaluates both internal and external evidence, and attempts to reconcile discrepancies. If reconciliation is not possible, the model is encouraged to transparently communicate residual uncertainty.

\subsection{CARE-RAG Generation.}
The above steps produce a conflict report through conflict-driven summarization, which effectively helps LLMs mitigate hallucinations caused by conflicting evidence. Finally, CARE-RAG feeds the parameter-aware evidence, context-aware evidence, and the corresponding conflict report into the LLM, enabling it to synthesize a final answer by reconciling conflicts and integrating all information. This enhances the transparency of parametric knowledge, factual accuracy, and robustness to conflicting or ambiguous evidence in the generated output.

\section{QA Repair for Valid Evaluation}



Standard QA benchmarks often suffer from outdated or mismatched ground truths, which can lead to inaccurate evaluations. 
Specifically, we conduct a manual analysis of 1,000 randomly sampled instances from each dataset and identify significant annotation flaws, as shown in Table~\ref{tab:Noise}. For instance, all 67 errors (100\%) in the Wiki dataset were due to outdated answers, while 44.6\% of the 74 errors in TriviaQA stemmed from semantic mismatches.

\begin{table}[t!]
\centering
\setlength{\tabcolsep}{4pt}
\begin{tabular}{lccc}
\toprule
\multirow{2}{*}{\textbf{Dataset}} & \multirow{2}{*}{\textbf{Repair}} &
\multicolumn{2}{c}{\textbf{Noise ratio (\%)}}  \\
\cmidrule(lr){3-4}
& & Mismatch & Outdate\\
\midrule
Wiki      & 67  & 0.0  & 100.0 \\

TriviaQA  & 74  & 44.6 & 55.4 \\

NQ        & 240 & 19.6 & 81.7 \\

HotpotQA  & 103 &  8.7 & 91.3 \\

ASQA      & 157 &  0.6 & 99.4 \\
\bottomrule
\end{tabular}
\caption{Prevalence of outdated or mismatched ground truths in standard QA benchmarks. Noise classification is based on manual analysis and repair of 1,000 sampled instances per dataset.}
\label{tab:Noise}
\end{table}

To address this issue, we introduce a QA Repair pre-processing step to ensure fairer comparisons.
For instance, on TriviaQA, this approach raises the F1 score from 85.09 to 86.17 for the Qwen3-235B-A22B model, as shown in Table~\ref{tab:union_all}. 
Further implementation details are provided in Appendix~\ref{app:qarepair}.






\begin{table}[ht]
\centering
\renewcommand{\arraystretch}{1.25}
\resizebox{\columnwidth}{!}{ 
\begin{tabular}{lcccc}
\toprule
\multirow{2}{*}{\textbf{Dataset}}  & Baseline & Mismatch & Outdate & Both \\
\cmidrule(lr){2-2} \cmidrule(lr){3-3} \cmidrule(lr){4-4} \cmidrule(lr){5-5}
& EM / F1 & EM / F1 & EM / F1 & EM / F1 \\
\midrule
Wiki      & 54.8 / 55.4 & 54.8 / 55.4 & 56.7 / 57.4 & \textbf{56.7 / 57.4} \\
TriviaQA  & 84.9 / 85.1 & 85.6 / 85.7 & 85.3 / 85.5 & \textbf{85.9 / 86.2} \\
NQ        & 71.2 / 71.5 & 72.4 / 72.8 & 75.5 / 75.9 & \textbf{76.0 / 76.3} \\
HotpotQA  & 63.1 / 63.6 & 63.9 / 64.3 & 66.8 / 67.3 & \textbf{67.1 / 67.5} \\
ASQA      & 59.8 / 60.1 & 60.1 / 60.4 & 62.6 / 63.1 & \textbf{62.9 / 63.3} \\
\bottomrule
\end{tabular}
}
\caption{QA performance improvements via QA Repair across datasets."Baseline" shows original scores; "Mismatch", "Outdate", and "Both" indicate results after fixing semantic mismatches, outdated answers, and both, respectively. All values are reported as EM/F1.}
\label{tab:union_all}
\end{table}


\section{Experimental Setup}
\label{sec:exp_framework}


\subsection{Datasets}
\label{sec:datasets_and_repair}

Our experimental evaluation utilizes five challenging QA benchmarks: Natural Questions (NQ)~\citep{kwiatkowski2019natural}, TriviaQA~\citep{joshi2017triviaqa}, HotpotQA~\citep{yang2018hotpotqa}, ASQA~\citep{stelmakh2022asqa}, and 2WikiMultiHopQA~\citep{zhang2023end}.
To ensure fair evaluation, we apply our QA Repair procedure to all five datasets, resulting in improved versions denoted as \(\text{NQ}^*\), \(\text{TriviaQA}^*\), \(\text{HotpotQA}^*\), \(\text{ASQA}^*\) and \(\text{WikiQA}^*\), which are used consistently throughout our experiments.
This process addresses common issues such as outdated or mismatched, enhancing alignment between model predictions and acceptable references.

\begin{table*}[ht]
  \noindent
   
  \resizebox{\textwidth}{!}{%
    \setlength{\tabcolsep}{7.5pt}
    \renewcommand{\arraystretch}{0.95}
    \footnotesize

    \begin{tabular}{l|cc|cc|cc|cc|cc}
      \toprule
      \multirow{2}{*}{\textbf{Method}}
      & \multicolumn{2}{c|}{\(\text{NQ}^*\)}
      & \multicolumn{2}{c|}{ \(\text{TriviaQA}^*\)}
      & \multicolumn{2}{c|}{ \(\text{HotpotQA}^*\)}
      & \multicolumn{2}{c|}{\(\text{ASQA}^*\)}
      & \multicolumn{2}{c}{\(\text{WikiQA}^*\)} \\
      & EM & F1 & EM & F1 & EM & F1 & EM & F1 & EM & F1 \\
      \midrule
      \multicolumn{11}{c}{\itshape Mistral-7B-v0.3} \\
      \midrule
      No RAG        & 39.7 & 41.6 & 65.2 & 66.8 & 35.8 & 38.5 & 32.3 & 34.6 & 33.2 & 36.9 \\
      RAG        & 41.4 & 42.7 & 66.0 & 67.2 & 34.7 & 36.4 & 32.2 & 34.2 & 35.9 & 37.7 \\
      InstructRAG   & 60.4 & 61.9 & 75.3 & 76.6 & 49.4 & 52.2 & 47.0 & 48.75 & 43.9 & 44.9 \\
      GenRead       & 48.9 & 49.3 & 70.7 & 71.0 & 38.6 & 39.3 & 37.8 & 38.3 & 37.7 & 38.5 \\
      Self-RAG      & 43.1 & 44.2 & 66.9 & 67.7 & 39.2 & 40.9 & 36.0 & 37.37 & 38.8 & 40.5 \\
      \textbf{CARE-RAG}
                    & \textbf{63.1} & \textbf{63.5}
                    & \textbf{78.4} & \textbf{78.8}
                    & \textbf{53.1} & \textbf{53.8}
                    & \textbf{50.6} & \textbf{51.1}
                    & \textbf{44.7} & \textbf{45.6} \\
      \midrule
      \multicolumn{11}{c}{\itshape Llama-3.2-8B} \\
      \midrule
      No RAG        & 39.9 & 42.4 & 64.6 & 67.13 & 32.6 & 36.1 & 32.6 & 36.1 & 33.7 & 40.2 \\
      RAG        & 40.3 & 42.5 & 66.1 & 68.4 & 35.3 & 39.1 & 33.2 & 36.4 & 33.9 & 39.3 \\
      InstructRAG   & 59.7 & 60.9 & 73.9 & 75.1 & 48.5 & 50.5 & 45.9 & 47.2 & 36.9 & 40.6 \\
      GenRead       & 50.9 & 51.2 & 73.5 & 73.9 & 40.5 & 41.4 & 40.9 & 41.6 & 38.1 & 39.5 \\
      Self-RAG& 40.8 & 42.5 & 68.3 & 70.2 & 36.9 & 39.9 & 34.2 & 36.9 & 34.2 & 39.1 \\
      \textbf{CARE-RAG}
                    & \textbf{63.9} & \textbf{64.3}
                    & \textbf{79.6} & \textbf{79.9}
                    & \textbf{55.9} & \textbf{56.6}
                    & \textbf{52.6} & \textbf{53.1}
                    & \textbf{47.1} & \textbf{48.0} \\
      \midrule
      \multicolumn{11}{c}{\itshape Qwen2.5-7B} \\
      \midrule
      No RAG        & 28.2 & 31.0 & 51.2 & 53.1 & 31.2 & 34.5 & 17.9 & 21.5 & 31.3 & 37.0 \\
      RAG        & 31.0 & 32.8 & 52.9 & 54.4 & 30.5 & 32.9 & 18.9 & 21.7 & 30.6 & 32.1 \\
      InstructRAG   & 60.7 & 61.3 & 72.7 & 74.3 & 52.4 & 53.6 & 47.7 & 48.5 & 39.8 & 41.3 \\
      GenRead       & 39.5 & 39.9 & 59.2 & 59.6 & 34.1 & 34.8 & 24.3 & 25.0 & 31.5 & 32.4 \\
      Self-RAG& 32.8 & 33.9 & 54.1 & 55.1 & 33.8 & 35.2 & 20.0 & 21.7 & 32.9 & 34.8 \\
      \textbf{CARE-RAG}
                    & \textbf{62.2} & \textbf{62.2}
                    & \textbf{75.4} & \textbf{75.7}
                    & \textbf{54.0} & \textbf{54.6}
                    & \textbf{50.8} & \textbf{51.3}
                    & \textbf{42.9} & \textbf{43.8} \\
      \bottomrule
    \end{tabular}
  }
    \caption{Comparing Conflict-Aware and Reliable Evidence for RAG with open-source models on five QA benchmarks (EM/F1 scores). CARE-RAG achieves superior performance across all datasets and models.}
  \label{open-source}
\end{table*}

\subsection{Implementation Details}
\label{sec:implementation_details_main}

We evaluate CARE-RAG using both open-source and closed-source LLMs.
The open-source models include Mistral-7B~\citep{jiang2023mistral7b}, LLaMA-3.2-8B~\citep{grattafiori2024llama}, and Qwen2.5-7B~\citep{yang2024qwen2}.
The closed-source models include Claude-3.5-Haiku~\citep{anthropic2024haiku}, Gemini-2.0-Flash~\citep{balestri2025gender}, and GPT-4.1-Nano~\citep{openai2025gpt41nano}.
Experiments use consistent hyperparameters across models (max\_tokens=1024, temperature=0.7). Inference for open-source models is conducted using VLLM~\citep{kwon2023efficient}, while closed-source models are accessed via official APIs.

We retrieve the top-5 most relevant evidences for each query, with retrieval sensitivity analysis (varying top-K from 5 to 25) reported in Section~\ref{sec:results_retrieval_robustness} and Appendix~\ref{app:implementation}.
Conflict Detection is powered by a distilled LLaMA-3.2B model fine-tuned on DeepSeek annotations, enabling efficient semantic conflict analysis.
parameter evidence ($\mathcal{E}_p$) is generated via iterative prompting to elicit diverse internal perspectives from the LLM.
Context refinement is guided by instruction-based prompting, with prompt templates detailed in Appendix~\ref{app:prompts}.

\subsection{Baselines}
\label{sec:baselines}

We compare CARE-RAG with four representative baselines, covering key paradigms in retrieval-augmented generation. \textbf{No RAG} uses only the LLM’s parameter knowledge, without any retrieved context, serving as a lower bound that reflects the limitations of internal knowledge alone. \textbf{InstructRAG}~\citep{wei2024instructrag} improves answer quality by prompting the LLM with rationale-based instructions over retrieved evidences, but lacks mechanisms to handle contradictions across evidence. \textbf{GenRead}~\citep{yu2022generate} compresses retrieved content into concise summaries before generation, mitigating retrieval noise but potentially omitting important conflicting signals. \textbf{Self-RAG}~\citep{asai2023self} incorporates a self-reflection stage to critique initial answers and refine retrieval, but does not explicitly model conflicts between internal and external knowledge. These baselines highlight the challenges of retrieval quality, hallucination, and inconsistency, which CARE-RAG addresses through structured introspection and conflict resolution.

\begin{table}[ht]
  \centering
  \small
  
  \renewcommand{\arraystretch}{1.0}
  \setlength{\tabcolsep}{4.0pt} 
  \begin{tabular}{l|cc|cc|cc}
    \toprule
    \multirow{2}{*}{\textbf{Method}} & \multicolumn{2}{c|}{\(\text{NQ}^*\)} & \multicolumn{2}{c|}{ \(\text{TriviaQA}^*\)} & \multicolumn{2}{c}{\(\text{WikiQA}^*\).} \\
    & EM & F1 & EM & F1 & EM & F1 \\
    \midrule
    \multicolumn{7}{c}{\itshape LLaMA-3.2-8B} \\
    \midrule
    w/o Stage1 & 61.5 & 62.8 & 77.3 & 72.61 & 43.3 & 44.7 \\
    w/o Stage2 & 39.9 & 42.44 & 64.6 & 67.13 & 33.7 & 40.17 \\
    w/o Stage3 & 60.3 & 60.74 & 78.12 & 77.93 & 44.1 & 45.29 \\
    \textbf{CARE-RAG} & \textbf{63.9} & \textbf{64.31} & \textbf{79.6} & \textbf{79.89} & \textbf{47.1} & \textbf{48.0} \\
    \midrule
    \multicolumn{7}{c}{\itshape Mistral-7B-v0.3} \\
    \midrule
    w/o Stage1 & 60.6 & 61.2 & 77.5 & 77.7 & 44.0 & 45.0 \\
    w/o Stage2 & 39.7 & 41.61 & 65.2 & 66.78 & 33.2 & 36.94 \\
    w/o Stage3 & 59.4 & 59.95 & 77.8 & 78.1 & 43.9 & 44.85 \\
    \textbf{CARE-RAG} & \textbf{63.1} & \textbf{63.54} & \textbf{78.4} & \textbf{78.8} & \textbf{44.7} & \textbf{45.61} \\
    \bottomrule
  \end{tabular}
  \caption{Ablation study showing that each component of CARE-RAG contributes to performance across \(\text{NQ}^*\), \(\text{TriviaQA}^*\), and \(\text{WikiQA}^*\) datasets.}
  \label{tab:reasoning_stages}
\end{table}

\section{Results and Analysis}
\label{sec:results_and_analysis}

\subsection{Overall Performance}
\label{sec:results_performance_analysis}

We evaluate CARE-RAG on five QA benchmarks (NQ*, TriviaQA*, HotpotQA*, ASQA*, WikiQA*) under both open-source and closed-source model settings, as shown in Tables~\ref{open-source} and \ref{close-source}. CARE-RAG consistently achieves the highest EM and F1 scores across all datasets and models. Compared to the standard RAG baseline, CARE-RAG improves performance by up to 17.2 EM and 17.1 F1. Relative to the strongest baseline method (InstructRAG), it still achieves an average improvement of 3.8 EM and 3.7 F1. Although closed-source models generally exhibit higher absolute performance due to larger scale and better pretraining, CARE-RAG maintains consistent gains in both open-source and closed-source settings, demonstrating its robustness and general applicability.

These results indicate that the core mechanisms of CARE-RAG—structured parameter introspection, evidence refinement, and conflict-aware summarization—are highly effective in enhancing answer reliability. By explicitly detecting and resolving contradictions between internal and retrieved knowledge, CARE-RAG improves factual accuracy without relying on handcrafted prompts or answer-level self-reflection. This architecture is particularly beneficial in scenarios involving noisy or conflicting evidence, where traditional RAG methods tend to fail. The consistent improvements across datasets and models support CARE-RAG’s potential as a general framework for trustworthy retrieval-augmented generation.

\begin{table*}[ht]
  \noindent
    
  \makebox[\linewidth][l]{%
    \resizebox{\linewidth}{!}{%
      \setlength{\tabcolsep}{7.5pt}
      \renewcommand{\arraystretch}{0.95}
      \footnotesize
      \begin{tabular}{l|cc|cc|cc|cc|cc}
        \toprule
        \multirow{2}{*}{\textbf{Method}}
        & \multicolumn{2}{c|}{\(\text{NQ}^*\)}
        & \multicolumn{2}{c|}{ \(\text{TriviaQA}^*\)}
        & \multicolumn{2}{c|}{ \(\text{HotpotQA}^*\)}
        & \multicolumn{2}{c|}{\(\text{ASQA}^*\)}
        & \multicolumn{2}{c}{\(\text{WikiQA}^*\).} \\
        & EM & F1 & EM & F1 & EM & F1 & EM & F1 & EM & F1 \\
        \midrule
        \multicolumn{11}{c}{\itshape claude-3-5-haiku-latest} \\
        \midrule
        No RAG     & 50.6 & 52.3 & 77.8  & 78.9 & 42.1 & 44.8 & 40.9 & 43.3 & 35.5 & 39.2 \\
        RAG        & 51.9 & 53.0 & 78.7 & 79.1 & 43.3 & 44.7 & 41.0 & 42.9 & 35.5 & 37.9 \\
        InstructRAG& 67.7 & 68.3 & 79.2  & 79.7 & 53.2 & 54.4 & 50.1 & 50.7 & 39.5 & 41.6 \\
        GenRead    & 57.0 & 57.5 & 80.1 & 80.4 & 43.9 & 44.7 & 46.3 & 47.0 & 35.4 & 36.5 \\
        Self-RAG& 52.9 & 53.8 & 79.1  & 79.8 & 44.9 & 46.6 & 41.1 & 42.56 & 36.8 & 39.0 \\
        \textbf{CARE-RAG}
                   & \textbf{68.8} & \textbf{69.2}
                   & \textbf{85.9} & \textbf{86.1}
                   & \textbf{57.9} & \textbf{58.6}
                   & \textbf{58.8} & \textbf{59.3}
                   & \textbf{47.5} & \textbf{48.3} \\
        \midrule
        \multicolumn{11}{c}{\itshape gemini-2.0-flash} \\
        \midrule
        No RAG     & 42.4 & 50.1 & 70.8  & 73.7 & 39.6 & 47.7 & 45.2 & 54.1 & 28.0 & 39.2 \\
        RAG        & 46.1 & 51.4 & 72.4 & 75.8 & 39.5 & 45.2 & 44.0 & 47.2 & 31.4 & 38.6 \\
        InstructRAG& 65.3 & 66.7 & 75.1  & 76.5 & 49.1 & 50.9 & 46.9 & 48.6 & 41.2 & 44.7 \\
        GenRead    & 57.5 & 57.9 & 82.6  & 83.9 & 48.7 & 49.3 & 49.6 & 49.7 & 44.4 & 45.2 \\
        Self-RAG& 49.4 & 52.4 & 77.5  & 78.7 & 39.4 & 41.8 & 42.6 & 45.3 & 34.5 & 38.0 \\
        \textbf{CARE-RAG}
                   & \textbf{68.0} & \textbf{68.5}
                   & \textbf{86.7} & \textbf{87.1}
                   & \textbf{61.4} & \textbf{62.3}
                   & \textbf{63.6} & \textbf{64.2}
                   & \textbf{56.7} & \textbf{57.7} \\
        \midrule
        \multicolumn{11}{c}{\itshape gpt-4.1-nano-2025-04-14} \\
        \midrule
        No RAG     & 35.8 & 40.0 & 62.0  & 64.8 & 31.7 & 37.6 & 28.0 & 33.0 & 31.1 & 39.4 \\
        RAG        & 39.4 & 43.6 & 65.4 & 67.7 & 33.9 & 38.3 & 31.7 & 35.2 & 31.8 & 39.0 \\
        InstructRAG& 58.5 & 60.5 & 72.5  & 73.6 & 53.5 & 56.24 & 48.1 & 50.08 & 40.4 & 44.5 \\
        GenRead    & 51.0 & 51.9 & 72.9  & 73.5 & 42.6 & 43.8 & 39.4 & 40.7 & 37.4 & 39.3 \\
        Self-RAG& 43.7 & 46.0 & 68.1  & 69.6 & 35.9 & 39.3 & 34.2 & 37.5 & 32.2 & 38.2 \\
        \textbf{CARE-RAG}
                   & \textbf{66.2} & \textbf{66.5}
                   & \textbf{81.6} & \textbf{81.2}
                   & \textbf{56.7} & \textbf{57.2}
                   & \textbf{53.0} & \textbf{53.4}
                   & \textbf{47.6} & \textbf{48.2} \\
        \bottomrule
      \end{tabular}
      
    }%
  }
  \caption{Comparing Conflict-Aware and Reliable Evidence for RAG with closed-source models on five QA benchmarks (EM/F1 scores). CARE-RAG achieves superior performance across all datasets and models.}
  \label{close-source}
\end{table*}

\subsection{Ablation Study of Core Components}
\label{sec:ablation}

To evaluate the effectiveness of CARE-RAG’s core components, we perform an ablation study under three settings. 
\textbf{w/o Stage1}: Removes the parameter Record Comparison stage and relies only on external retrieved evidence for answer generation.
\textbf{w/o Stage2}: Removes the Retrieval Result Refinement module.
\textbf{w/o Stage3}: Removes the Conflict-Driven Summarization stage, omitting both conflict.

As shown in Table~\ref{tab:reasoning_stages}, all three components contribute significantly to the overall performance of CARE-RAG across datasets and model backbones.
1) Introducing external retrieved evidence and refining it into a structured Context-aware evidence (\(\mathcal{E}_c\)) leads to substantial gains over using parameter knowledge alone. For example, on the NQ dataset with \textsc{LLaMA}-3-8B, adding refined external evidence yields a +20.4 EM improvement. This highlights the importance of incorporating external information in a structured and relevant form via \(\Pi_{\text{ref}}\); 
2) Adding explicit conflict resolution—through conflict detection (\(\mathcal{M}_c\)) and conflict-aware answer synthesis (\(\Pi_{\text{synth}}\))—provides consistent additional gains of 1--2 EM/F1. This shows the value of not only using external knowledge but also explicitly identifying and reconciling inconsistencies between the internal parameter knowledge (\(\mathcal{E}_p\)) and the retrieved evidence (\(\mathcal{E}_c\)). Such targeted conflict handling is crucial for ensuring factual consistency and generating trustworthy answers, leading to the full performance of CARE-RAG.

\subsection{Sensitivity to Retrieval Volume} 
\label{app:sensitivity_topk} 

\begin{figure}[htbp]
  \centering

  \subfigure[\(\text{NQ}^*\)/EM]{
    \includegraphics[width=0.46\linewidth]{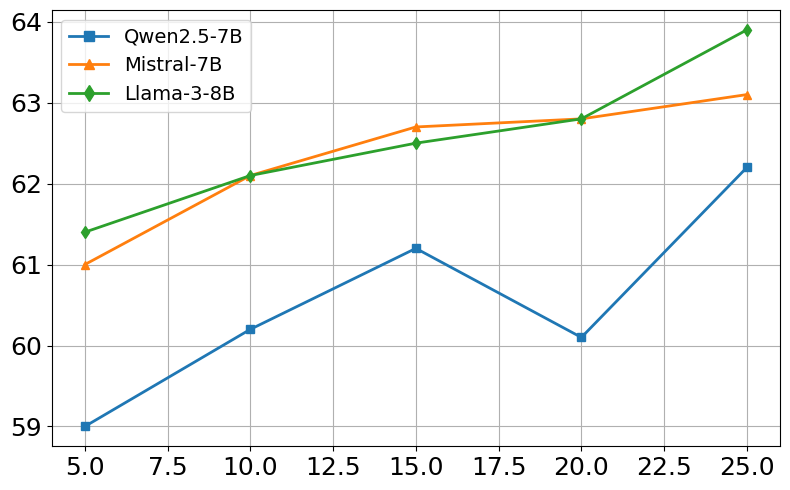}
  }
  \subfigure[\(\text{NQ}^*\)/F1]{
    \includegraphics[width=0.46\linewidth]{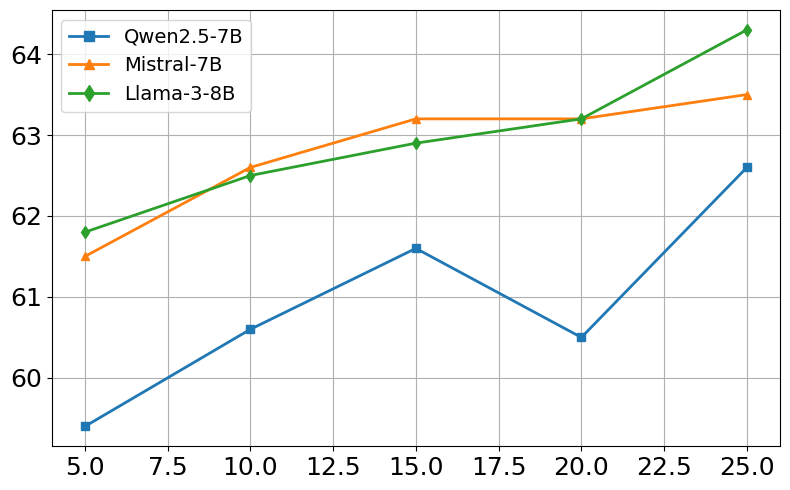}
  }

  \subfigure[ \(\text{HotpotQA}^*\)/EM]{
    \includegraphics[width=0.46\linewidth]{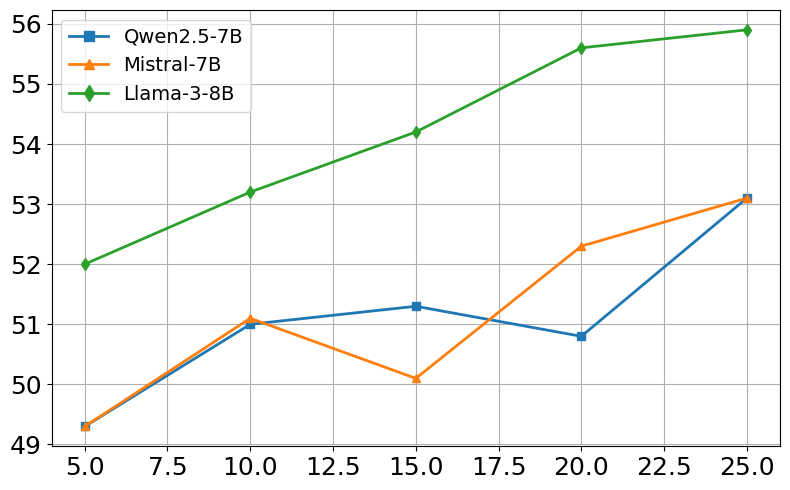}
  }
  \subfigure[ \(\text{HotpotQA}^*\)/F1]{
    \includegraphics[width=0.46\linewidth]{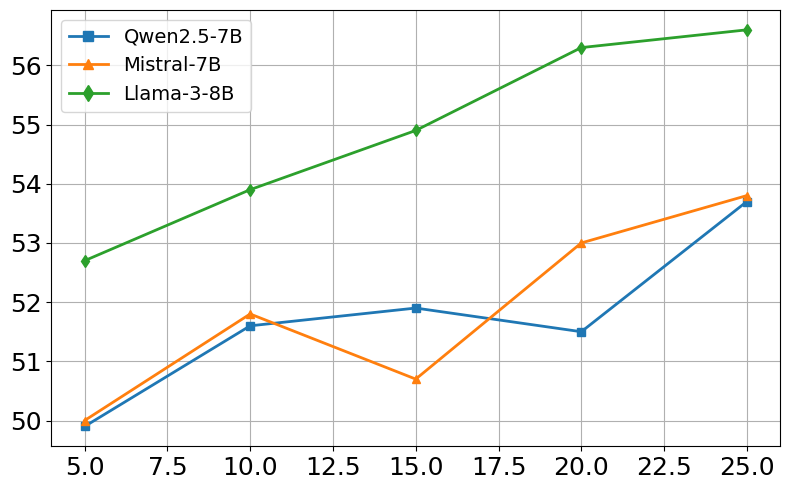}
  }

  \caption{Sensitivity to retrieval size ($K$). EM/F1 scores for NQ and HotpotQA across three open-source models.}
  \label{fig:fig_3}
\end{figure}

To assess the robustness of our method under varying retrieval volumes, we conduct an ablation study on $K$, the number of retrieved evidences. We evaluate CARE-RAG under different retrieval volumes, varying $K$ from 5 to 25. The results are presented in Figure~\ref{fig:fig_3}.

The findings indicate that CARE-RAG effectively utilizes increased context, benefiting particularly from its context refinement mechanism $\Pi_{\text{ref}}$. Performance generally peaks around $K=15\text{--}20$, beyond which it plateaus, showing remarkable stability even when potentially lower-quality or redundant evidence is included. This contrasts with simpler RAG methods, which often suffer from noise accumulation at higher $K$ values. These results suggest that CARE-RAG's structured reasoning and conflict resolution mechanisms are effective at filtering and prioritizing information, thereby maintaining performance even under noisy conditions.

    
    

\subsection{Robustness to Retrieval Variations}
\label{sec:results_retrieval_robustness}

\begin{figure}[!t]
  \centering
  \includegraphics[width=\columnwidth]{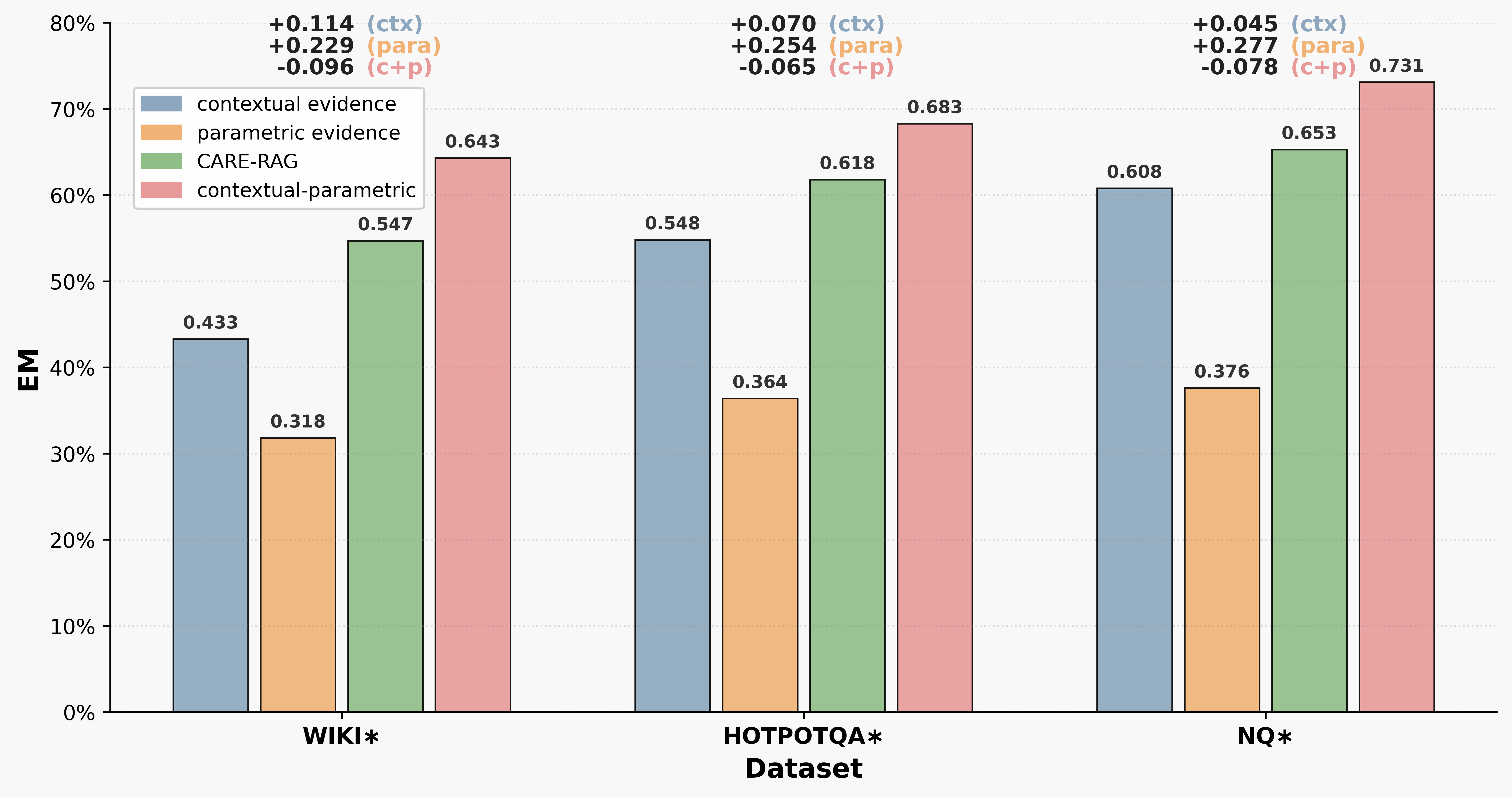}
  \caption{EM performance across three datasets using different retrieval evidence sources.
}
  \label{fig:retrieval_quality}
\end{figure}

A robust RAG system must remain effective under imperfect retrieval conditions, where the provided evidence may vary significantly in relevance, completeness, or even contradict the original query intent. Figure~\ref{fig:retrieval_quality} illustrates EM scores across three datasets under four different evidence strategies: contextual only, parameter only, their direct combination, and CARE-RAG.  
CARE-RAG consistently outperforms both contextual and parameter-only baselines, achieving gains of up to 0.239 EM. These trends highlight CARE-RAG’s superior robustness to variations in evidence quality and composition, especially in scenarios with conflicting or incomplete information.

This robustness stems from CARE-RAG’s conflict-aware synthesis process: contextual evidence $\mathcal{E}_c$, retrieved and refined through $\Pi_{\text{ref}}$, is systematically compared with parameter-derived knowledge $\mathcal{E}_p$ using the conflict detector $\mathcal{M}_c$. This enables the model to identify and suppress misleading or contradictory signals, prioritize reliable content, and ultimately produce more accurate and trustworthy answers even in noisy or adversarial retrieval settings.


\section{Related Work}
\label{sec:related_work}

RAG aims to enhance Large Language Models (LLMs) by incorporating external knowledge~\citep{lewis2020retrieval, guu2020retrieval}. Early work and pretraining objectives focused on effective retrieval integration~\citep{izacard2023atlas}. However, RAG's reliability remains challenged by the quality of retrieved information and the model's ability to integrate it with internal knowledge. Improving the retriever itself is an active research area, with utility-based methods like shared context attribution enhancing relevance and usefulness of evidences~\citep{xu2025training}.

Beyond basic retrieval and integration, several methods aim to improve RAG systems. For instance, REPLUG~\citep{shi2023replug} explored black-box retrieval integration, while RA-DIT~\citep{lin2023ra} and InstructRetro~\citep{wang2023instructretro} investigated instruction tuning for better downstream task alignment. RankRAG~\citep{yu2024rankrag} improved passage ranking. Yet, these approaches may not fully resolve challenges from conflicting or low-quality retrievals, or the critical issue of maintaining generation faithfulness—addressed by works like~\citep{bi2024context,zhang2025faithfulrag}, which align LLMs for context-faithful outputs. Our proposed Conflict-Aware and Reliable Evidence for RAG (CARE-RAG) explicitly targets post-retrieval synthesis to improve robustness under such scenarios.

A core challenge in RAG lies in resolving knowledge conflicts—when retrieved content contradicts either the LLM's prior or other evidences~\citep{wang2023instructretro, zhou2025trustrag, zou2024poisonedrag, jin2024tug, xie2023adaptive, bi2025parameters}. These conflicts can cause factual inaccuracies and intersect with the domain of knowledge editing in LLMs, where methods seek to correct or bias internal representations~\citep{li2025reinforced, zhang2025explainable} or reinforce edited knowledge through contrastive decoding~\citep{bi2024decoding}. Such concerns resonate with findings on data noise sensitivity~\citep{jiang2024instruction, chen2024benchmarking}, and underscore the need for conflict resolution within RAG pipelines. This is especially relevant in adversarial or noisy scenarios, including graph-based settings where even single-node attacks can distort outcomes~\citep{DBLP:journals/corr/abs-2108-13049}. Ensuring fair evaluation thus requires accounting for both retrieval quality and model alignment~\citep{jacovi2023stop}.

\section{Conclusion}
\label{sec:conclusion}

CARE-RAG is a conflict-aware and reliable framework for retrieval-augmented question answering that systematically tackles key reliability challenges, including outdated supervision, noisy or conflicting retrievals, and inconsistencies between internal and external knowledge. By integrating structured parameter introspection, fine-grained context refinement, lightweight conflict detection, and a QA repair mechanism, CARE-RAG enhances factual consistency and robustness across diverse QA tasks. Extensive experiments on five benchmarks and multiple model backbones show that CARE-RAG consistently outperforms strong baselines, underscoring the value of explicitly modeling knowledge conflicts for trustworthy and generalizable retrieval-augmented generation.

\section*{Limitations}
CARE-RAG demonstrates notable improvements over existing retrieval-augmented methods; however, certain limitations remain. The multi-stage approach inherently incurs greater computational overhead compared to simpler RAG frameworks, potentially impacting inference efficiency. Additionally, the performance of CARE-RAG, particularly its conflict detection and resolution capabilities, remains closely tied to the quality of the underlying language models and their fine-tuned capabilities, which might not fully resolve highly subtle or adversarially constructed knowledge conflicts. Furthermore, despite increased robustness to noisy retrieval, overall efficacy still depends substantially on the initial document retriever’s accuracy and comprehensiveness. The QA Repair module, though effective against typical dataset issues, may not universally handle all types of benchmark artifacts or specialized domain knowledge without further refinement and domain-specific adaptation.

\section*{Ethical Considerations}
\label{sec:ethical_considerations}
The development of advanced retrieval-augmented generation systems, including CARE-RAG, raises significant ethical considerations. The QA Repair process, designed to address dataset biases by correcting outdated or mismatched information, inherently involves subjective judgments regarding the definition and scope of "correctness." Such judgments must be transparently managed and periodically revisited to prevent inadvertent bias introduction. Additionally, improvements in factual accuracy and consistency, although broadly beneficial, increase the risk of generating convincing yet inaccurate information if misused or inadequately supervised. Reliance on externally retrieved knowledge also introduces the possibility of propagating existing biases or inaccuracies from source materials. Therefore, ongoing research efforts should emphasize robust bias detection, clear attribution of information sources, transparent conflict-resolution mechanisms, and the establishment of responsible use guidelines to ensure these powerful tools are deployed ethically, fairly, and constructively.

\bibliography{custom}
\newpage
\appendix

\section{Implementation Details}
\label{app:implementation}

\subsection{Models Used}
\label{app:impl:models}
Our experiments primarily utilized three open-source Large Language Models (LLMs): Mistral-7B~\citep{jiang2023mistral7b}, Llama-3.2-8B~\citep{grattafiori2024llama}, and Qwen2.5-7B~\citep{yang2024qwen2}. For experiments involving closed-source models (as detailed in Table~\ref{close-source}), we employed. 

Unless otherwise specified, the same backbone LLM (from either the open-source or closed-source set, depending on the experiment) was consistently used for all stages of the CARE-RAG pipeline: eliciting the initial LLM response (\(A_{\text{init}}\), corresponding to \(\mathcal{E}_p\) generation in Algorithm~\ref{alg:srrag_detailed}), refining retrieved results into \(\mathcal{E}_c\) (which may involve structured reasoning), conflict detection by \(\mathcal{M}_c\) to produce \(\delta_c\) and \(r_c\), CARE-RAG to generate \(\hat{a}\), and also for the QA Repair module (\(f_{\text{repair}}\)) described in Appendix~\ref{app:qarepair}.

\subsection{Retrieval Setup}
\label{app:impl:retriever}

For each question, we retrieved the top-K relevant evidences from the respective corpus. In our main experiments (Tables~\ref{open-source} and \ref{close-source}), K was set to 5. An analysis of CARE-RAG's sensitivity to varying K (from 5 to 25) is presented in Figure~\ref{fig:fig_3}.

\subsection{Inference Framework}
\label{app:impl:inference}
All inferences for open-source LLMs were performed using the vLLM framework~\citep{kwon2023efficient} to ensure efficiency and reproducibility. For closed-source models, inferences were made via their respective official APIs.
The following inference parameters were consistently applied for generation tasks (e.g., generating initial responses for \(\mathcal{E}_p\), the refined evidence \(\mathcal{E}_c\), the final answer \(\hat{a}\), and repaired answers in \(f_{\text{repair}}\)) unless a specific module (like the conflict detector \(\mathcal{M}_c\)) required different settings: 
\begin{itemize}
    \item \texttt{max\_tokens}: 1024
    \item \texttt{temperature}: 0.7
    \item \texttt{top\_p}: 1.0
\end{itemize}
For classification-like tasks performed by the conflict detector \(\mathcal{M}_c\) to determine \(\delta_{c}\) (and generate \(r_c\)), we typically used a temperature (e.g., 0.7, or potentially lower like 0.0 for more deterministic conflict/no-conflict output if desired) to encourage more deterministic outputs, though the primary mechanism for binary classification was specific instruction prompting tailored to elicit a "0" or "1" and a rationale. 

\subsection{Evaluation Metrics}
\label{app:impl:metrics}
System performance was primarily evaluated using standard Exact Match (EM) and F1 scores. These metrics were computed against the (potentially) repaired ground truth answers generated by our QA Repair module (detailed in Appendix~\ref{app:qarepair}), ensuring a fair and robust assessment across all compared methods.

\section{QA Repair Module}
\label{app:qarepair}

\subsection{Overview}
\label{app:qarepair:overview}
As highlighted in our experimental setup (Section~\ref{sec:exp_framework}), standard QA benchmarks often suffer from issues such as temporal drift (outdated answers) or semantic mismatches between questions and ground truths. These flaws can lead to misleading evaluations of RAG systems. To ensure a fairer and more accurate assessment of model capabilities, we introduce a QA Repair module. This module is applied as a pre-processing step to the test instances of all evaluated benchmarks, correcting potential issues in the original ground truth answers before any model evaluation takes place. The module operates on an input triplet: (\textit{question \(q\)}, \textit{original ground truth answer} \(a_{\text{gt}}\), and potentially relevant \textit{retrieved context \(C\)}, though \(C\) is not always strictly necessary for the repair logic if general world knowledge suffices).

\subsection{Repair Mechanism}
\label{app:qarepair:mechanism}
The core of the QA Repair module is a classifier, \(f_{\text{repair}}\), implemented using a prompted Large Language Model (LLM). This classifier is tasked with assessing whether the original ground truth answer, \(a_{\text{gt}}\), is likely outdated, semantically inconsistent with the question \(q\), or otherwise flawed, considering current world knowledge and the precise intent of \(q\). It outputs a binary flag:
\[ \gamma_{\text{repair}} = f_{\text{repair}}(q, a_{\text{gt}}, C_{\text{optional}}) \in \{0, 1\} \]
If \(\gamma_{\text{repair}} = 1\) (indicating a detected flaw), a repair process is initiated. This process, also typically leveraging a prompted LLM, employs structured reasoning or direct knowledge querying (based on \(q\) and potentially \(C\)) to generate a revised, more accurate ground truth answer, \(a'_{\text{gt}}\). In some instances, to resolve ambiguity or align with the corrected answer, the original question \(q\) might also be minimally refined to \(q'\). The output of this stage is thus a potentially corrected question-answer pair \((q', a'_{\text{gt}})\). This repaired pair is then used as the reference for evaluating all RAG models (including baselines and CARE-RAG) in our experiments.

\subsection{Illustrative Examples}
\label{app:qarepair:examples}
The following examples illustrate typical scenarios handled by the QA Repair module. Note that in these examples, "Current Model Answer" (if such a term was used previously, otherwise this clarification might not be needed) is re-interpreted as the "Repaired Ground Truth (\(a'_{\text{gt}}\))" produced by our QA Repair module if a flaw was detected in the original "Ground Truth (\(a_{\text{gt}}\))".

\subsubsection{Example 1: Temporal Drift}
\begin{tcolorbox}[colframe=black!75, colback=white, sharp corners=south, boxrule=0.5pt, arc=3pt, fontupper=\small] 
\textbf{\textit{Scenario: Temporal Drift}}
\vspace{0.5em}

\textbf{Original Query (\(q\))}: Who scored the most points in their NBA career?
\vspace{0.5em}

\textbf{Original Ground Truth (\(a_{\text{gt}}\))}: Kareem Abdul-Jabbar
\vspace{0.5em}

\textbf{QA Repair Module Output}:
\begin{itemize}
    \item \textbf{Detection (\(\gamma_{\text{repair}}=1\))}: The answer "Kareem Abdul-Jabbar" is outdated.
    \item \textbf{Repaired Ground Truth (\(a'_{\text{gt}}\))}: \textcolor{blue}{LeBron James} (as of [current date/year of dataset repair])
    \item \textbf{Repaired Query (\(q'\))}: (No change in this case) Who scored the most points in their NBA career?
\end{itemize}
\end{tcolorbox}

\subsubsection{Example 2: Answer Type Mismatch / Factual Inaccuracy}
\begin{tcolorbox}[colframe=black!75, colback=white, sharp corners=south, boxrule=0.5pt, arc=3pt, fontupper=\small] 
\textbf{\textit{Scenario: Answer Type Mismatch / Factual Inaccuracy}}
\vspace{0.5em}

\textbf{Original Query (\(q\))}: When was the Statue of Liberty in France built?
\vspace{0.5em}

\textbf{Original Ground Truth (\(a_{\text{gt}}\))}: Paris
\vspace{0.5em}

\textbf{QA Repair Module Output}:
\begin{itemize}
    \item \textbf{Detection (\(\gamma_{\text{repair}}=1\))}: The answer "Paris" does not answer "When" and is factually incorrect for the construction date.
    \item \textbf{Repaired Ground Truth (\(a'_{\text{gt}}\))}: \textcolor{blue}{Construction was completed in July 1884.} (Or simply: \textcolor{blue}{July 1884})
    \item \textbf{Repaired Query (\(q'\))}: (No change in this case) When was the Statue of Liberty in France built?
\end{itemize}
\end{tcolorbox}

\subsubsection{Detailed Analysis of Repaired Data}
\label{app:qarepair:analysis} 

\begin{figure}[!ht] 
  \centering
  \includegraphics[width=0.9\columnwidth]{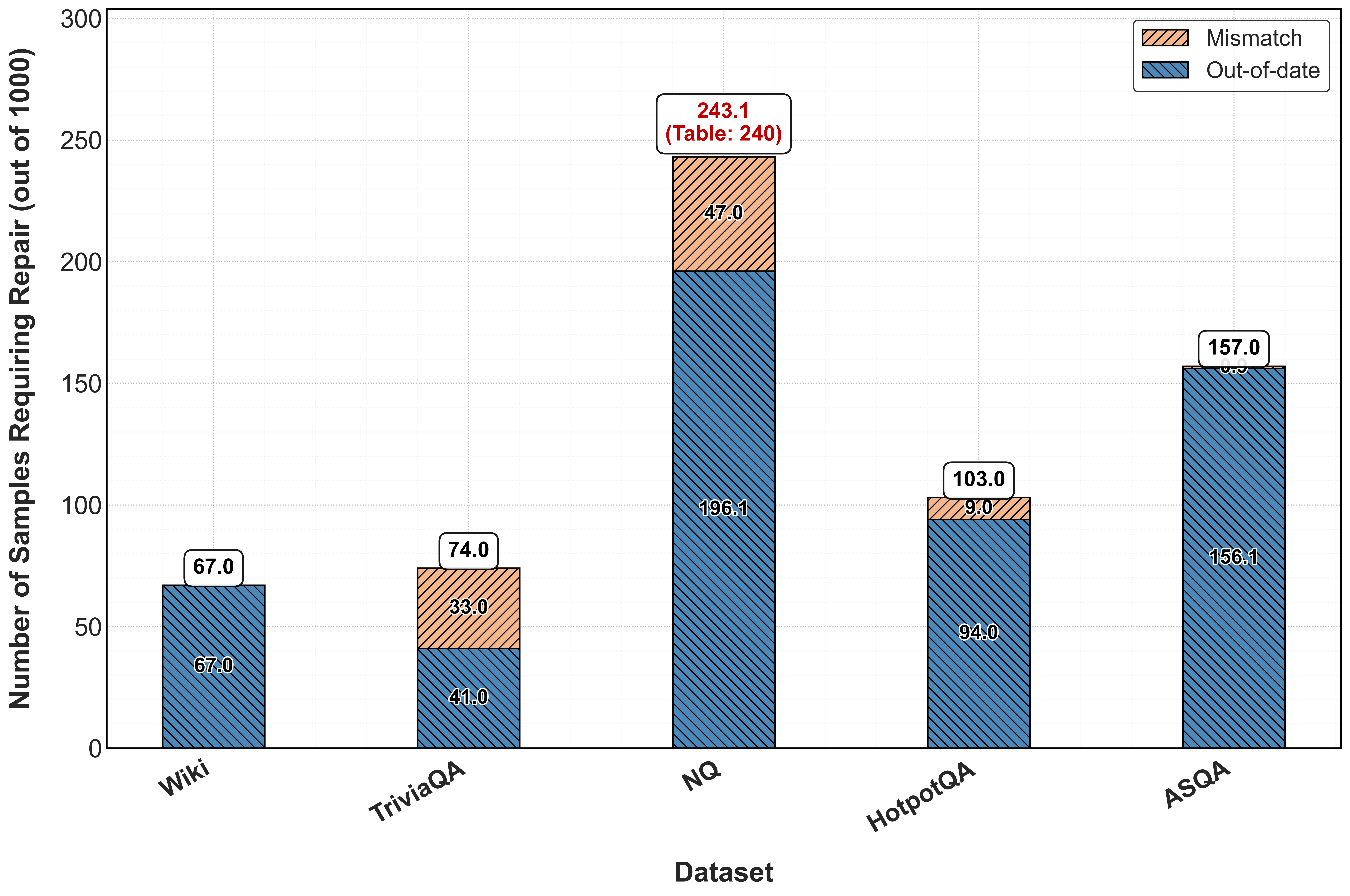} 
  \caption{Mismatch and Out-of-date error distribution in repaired samples from five QA datasets. NQ shows co-occurring error types. Repair impact on Qwen2.5-7B (using the notation from your paper if Qwen3-235B-A22B is a specific variant) is detailed in Table~\ref{tab:union_all} (Please verify this table label and its content regarding repair impact specifically with this model version).}
  \label{fig:repair_details_chart}
\end{figure}

Figure~\ref{fig:repair_details_chart} details the error composition within corrected samples from five QA benchmarks (1,000 samples each were analyzed for repair needs). The chart displays the counts of "Mismatch" errors (semantic misalignment) and "Out-of-date" errors (temporal drift) among the instances that required repair. For example, all 67 repaired Wiki samples were out-of-date, while TriviaQA's 74 repairs included approximately 33 mismatches. Notably, the NQ dataset, with 240 repaired samples, exhibits an overlap in error types: the sum of its reported mismatch (approx. 47) and out-of-date (approx. 196) components exceeds the total repair count, indicating some samples possess both error attributes. This granular analysis, highlighting diverse error profiles and potential co-occurrences as in NQ, underscores the necessity of our comprehensive QA Repair process for establishing a reliable evaluation baseline and the importance of targeted, rather than one-size-fits-all, approaches to dataset noise.
\section{Core Conflict-Aware and Reliable Evidence for RAG Prompts}
\label{app:prompts}

This section provides the specific prompt formats used for the core stages of the Conflict-Aware and Reliable Evidence for RAG (CARE-RAG) framework, corresponding to the \(\Pi\) symbols in Algorithm~\ref{alg:srrag_detailed}.

\subsection{Parameter Record Comparison Prompts (\texorpdfstring{$\Pi_{\text{init}}$ and $\Pi_{\text{iter}}$}{Pi_init and Pi_iter})}
\label{app:prompt:parameter_record}
\subsubsection{Iterative parameter Response Prompt (\texorpdfstring{$\Pi_{\text{iter}}$}{Pi_iter})}
\textit{Objective: Elicit alternative or more diverse parameter responses $a_i$ ($i > 0$), given the previously generated parameter responses within $\mathcal{E}_p$, to further explore the model's internal knowledge space.}

\begin{tcolorbox}[
  colframe=black!50, colback=gray!5, boxrule=0.5pt, arc=3pt, width=\linewidth,
  fontupper=\footnotesize, sharp corners=south, left=2mm, right=2mm, top=1mm, bottom=1mm
]
\textbf{\textcolor{black}{Task}}: Based on your previous answer(s) and your internal knowledge, provide a different or more detailed/nuanced answer to the following question.
\vspace{0.6em}

\textbf{\textcolor{blue!70!black}{Question}}: \{question\}
\vspace{0.6em}

\textbf{\textcolor{orange!80!black}{Previous parameter Answer(s) ($\mathcal{E}_p$ so far)}}: \{previous\_parameter\_answers\}
\vspace{0.6em}

\textbf{\textcolor{orange!80!black}{Answer (Iterative - $a_i$)}:}
\end{tcolorbox}
\subsubsection{Initial parameter Response Prompt (\texorpdfstring{$\Pi_{\text{init}}$}{Pi_init})}
\textit{Objective: Elicit the model's first direct response $a_0$ based solely on its internal parameter knowledge, forming the basis of \(\mathcal{E}_p\).}

\begin{tcolorbox}[
  colframe=black!50, colback=gray!5, boxrule=0.5pt, arc=3pt, width=\linewidth,
  fontupper=\footnotesize, sharp corners=south, left=2mm, right=2mm, top=1mm, bottom=1mm
]
\textbf{\textcolor{black}{Task}}: Provide a concise and direct answer to the following question using only your internal knowledge.
\vspace{0.6em}

\textbf{\textcolor{blue!70!black}{Question}}: \{question\}

\vspace{0.6em}
\textbf{\textcolor{orange!80!black}{Answer (Initial - $a_0$)}:}
\end{tcolorbox}

\subsection{Retrieval Result Refinement Prompt (\texorpdfstring{$\Pi_{\text{ref}}$}{Pi_ref})}
\label{app:prompt:retrieval_refinement}
\textit{Objective: Instruct the model to distill the retrieved evidences $C$ into a concise and salient context-aware evidence $\mathcal{E}_c$, by extracting key factual observations, identifying ambiguities, and forming context-grounded conclusions. This corresponds to Stage II in Figure~\ref{fig:framework}.}

\begin{tcolorbox}[
  colframe=black!30, colback=gray!5, boxrule=0.4pt, arc=2pt, width=\linewidth,
  fontupper=\footnotesize, sharp corners=south, left=2mm, right=2mm, top=1mm, bottom=1mm,
  before skip=6pt, after skip=6pt
]
\textbf{\textcolor{black}{Context Refinement Prompt}}
\vspace{0.5em}
\noindent \textbf{\textcolor{blue!70!black}{Instruction}}:
Analyze the provided Context thoroughly in relation to the Question. Your goal is to extract the most relevant factual information, identify any ambiguities or limitations within the context, and conclude with the most likely answer(s) or key insights that can be *purely grounded in the provided Context*. If no complete answer is available from the context, state that and explain why.
\vspace{0.5em}
\noindent \textbf{\textcolor{orange!80!black}{Retrieved Context evidences ($C$)}:}
\begin{itemize}[leftmargin=1.8em, itemsep=0.2em]
    \item \{context\_evidence\_1\}
    \item \{context\_evidence\_2\}
    \item ...
    \item \{context\_evidence\_k\}
\end{itemize}
\vspace{0.5em}
\noindent \textbf{\textcolor{teal!70!black}{Question ($q$)}:}
\{question\}
\vspace{0.5em}
\noindent \textbf{\textcolor{purple}{Your Distilled Context-Aware evidence ($\mathcal{E}_c$) based *only* on the Retrieved Context should include}}:
\begin{itemize}[leftmargin=1.8em, itemsep=0.2em]
    \item Key factual claims relevant to the Question.
    \item Identified ambiguities or limitations in the provided Context.
    \item A concluding summary or answer candidate(s) strictly derived from the Context.
\end{itemize}
\end{tcolorbox}

\subsection{Conflict Detection Prompt (\texorpdfstring{$\Pi_{c}$}{Pi_c})}
\label{app:prompt:conflict_detection}
\textit{Objective: Explicitly evaluate whether the model's consolidated parameter-aware evidences ($\mathcal{E}_p$) semantically conflict with the refined Context-aware evidence ($\mathcal{E}_c$). This is used by the conflict detector module $\mathcal{M}_c$ and corresponds to Stage III in Figure~\ref{fig:framework}.}

\begin{tcolorbox}[
  colframe=red!75,
  colback=red!5,
  boxrule=0.5pt, arc=3pt, width=\linewidth,
  fontupper=\footnotesize, sharp corners=south, left=2mm, right=2mm, top=1mm, bottom=1mm
]
\textbf{\textcolor{black}{Conflict Detection Prompt}}

\vspace{0.6em}
\textbf{\textcolor{blue!70!black}{Instruction}}: Evaluate if the "parameter Knowledge Response" contradicts the "Context-derived Response" for the given Question. Consider factual differences (e.g., names, dates, values), temporal mismatches, or significant semantic inconsistencies.
Output 'Conflict: 1' if a contradiction is found.
Output 'Conflict: 0' if there is no contradiction or if they are consistent.
Provide a brief step-by-step reasoning for your decision.

\vspace{0.6em}
\textbf{\textcolor{teal!70!black}{Question ($q$)}:} \{question\}

\vspace{0.6em}
\textbf{\textcolor{orange!80!black}{parameter Knowledge Response (Consolidated from $\mathcal{E}_p$)}:} \{consolidated\_parameter\_response\} 

\vspace{0.6em}
\textbf{\textcolor{purple}{Context-derived Response (from $\mathcal{E}_c$)}:} \{context\_aware\_evidence\_summary\} 

\vspace{0.6em}
\textbf{\textcolor{black}{Analysis and Conflict Decision ($\delta_c, r_c$)}:}
\end{tcolorbox}

\needspace{10\baselineskip} 
\subsection{CARE-RAG Generation Prompt (\texorpdfstring{$\Pi_{\text{synth}}$}{Pi_synth})}
\label{app:prompt:reflective_synthesis}
\textit{Objective: Generate the final answer ($\hat{a}$) by integrating the parameter-aware evidences ($\mathcal{E}_p$), the refined Context-aware evidence ($\mathcal{E}_c$), and the conflict detection signal ($\delta_c, r_c$). This corresponds to Stage IV in Figure~\ref{fig:framework}.}

\begin{tcolorbox}[
  colframe=black!30, colback=gray!5, boxrule=0.4pt, arc=2pt, width=\linewidth,
  fontupper=\footnotesize, sharp corners=south, left=2mm, right=2mm, top=1mm, bottom=1mm,
  before skip=6pt, after skip=6pt
]
\textbf{\textcolor{black}{Final Answer Synthesis Prompt}}
\vspace{0.5em}
\noindent \textbf{\textcolor{blue!70!black}{Contextual Note}}:
A potential conflict (indicated by \(\delta_c\)) between internal parameter knowledge (\(\mathcal{E}_p\)) and external information (\(\mathcal{E}_c\)) might have been detected, with rationale \(r_c\).
\vspace{0.5em}
\noindent \textbf{\textcolor{teal!70!black}{Your Task is to Synthesize the Best Final Answer ($\hat{a}$)}}:
\begin{enumerate}[leftmargin=1.8em, itemsep=0.2em]
    \item Based on all inputs, identify the \textbf{best-supported single candidate answer}.
    \item Consider information recency, source reliability, and overall coherence, especially if a conflict ($\delta_c=1$) was detected.
    \item \textbf{If conflict ($\delta_c=1$)}: Explicitly address the discrepancy from \(r_c\). Attempt to resolve it by selecting more credible information or state remaining uncertainty.
    \item \textbf{If no conflict ($\delta_c=0$)}: Primarily ground your answer in \(\mathcal{E}_c\), using \(\mathcal{E}_p\) as confirmation.
    \item Provide \textbf{concise reasoning} for your chosen answer, citing relevant inputs (\(\mathcal{E}_p, \mathcal{E}_c, r_c\)). Clearly state any remaining ambiguity or temporal uncertainty.
\end{enumerate}
\vspace{0.5em}
\noindent \textbf{\textcolor{orange!80!black}{Inputs Provided}}:
\begin{itemize}[leftmargin=1.8em, itemsep=0.2em]
    \item \textbf{Question ($q$)}: \{question\}
    \item \textbf{parameter Knowledge Response (Consolidated $\mathcal{E}_p$)}: \{consolidated\_parameter\_response\}
    \item \textbf{Context-derived Response ($\mathcal{E}_c$)}: \{context\_aware\_evidence\_summary\}
    \item \textbf{Conflict Detection Flag ($\delta_c$)}: \{\(\delta_c\)\}
    \item \textbf{Conflict Rationale ($r_c$)}: \{\(r_c\)\}
\end{itemize}
\vspace{0.5em}
\noindent \textbf{\textcolor{purple}{Required Output Format for Final Answer ($\hat{a}$)}}:
\begin{itemize}[leftmargin=1.8em, itemsep=0.2em]
    \item \textbf{Final Answer}: ...
    \item \textbf{Reasoning for Final Answer}: ... (Address conflict per \(r_c\) if $\delta_c=1$)
    \item \textbf{Ambiguity/Uncertainty Assessment}: ... (If any)
\end{itemize}
\end{tcolorbox}
\section{Detailed Process Walkthrough}
\label{app:walkthrough}

We illustrate the complete CARE-RAG workflow using the NBA scoring example, as discussed in Figure~\ref{fig:framework} (Stage I-IV visual overview) and referenced in the main text.

\begin{enumerate}[leftmargin=1.5em, itemsep=1em] 

\item \textbf{Input Question ($q$)}:
\textit{"Who scored the most points in their NBA career?"}

\item \textbf{parameter Record Comparison (generates $\mathcal{E}_p$)}:
The LLM \(\mathcal{M}\), using prompt \(\Pi_{\text{init}}\) (Appendix~\ref{app:prompt:parameter_record}), generates its initial context-free response \(a_0\). For this example, we assume \(n=1\), so the consolidated parameter-aware evidences \(\mathcal{E}_p\) is: \textit{"LeBron James"} (assuming the LLM's parameter knowledge is up-to-date).

\item \textbf{Retrieval Result Refinement (generates $\mathcal{E}_c$)}:
The retriever \(\mathcal{R}\) returns evidences \(C\), e.g.:
\(c_1\): \textit{"Kareem Abdul-Jabbar is the all-time leading scorer in the NBA, with 38,387 total points."};
\(c_2\): \textit{"Kareem rewriting scoring records."};
\(c_3\): \textit{"As of 2023, James holds the record."}
\newline
Using prompt \(\Pi_{\text{ref}}\) (Appendix~\ref{app:prompt:retrieval_refinement}), \(\mathcal{M}\) processes \(C\) into the Context-aware evidence \(\mathcal{E}_c\). For example, \(\mathcal{E}_c\) might be distilled to: \textit{"Retrieved evidences state Kareem Abdul-Jabbar was the all-time leading scorer (38,387 points). One passage indicates that as of 2023, James holds the record, suggesting a change."}

\item \textbf{Conflict-Driven Summarization (generates $\delta_c, r_c$)}:
The conflict detector \(\mathcal{M}_c\), using prompt \(\Pi_{c}\) (Appendix~\ref{app:prompt:conflict_detection}), compares \(\mathcal{E}_p\) (\textit{"LeBron James"}) with \(\mathcal{E}_c\) (\textit{"Retrieved evidences state Kareem... James holds the record..."}).
\newline
Assuming for clearer conflict demonstration that \(\mathcal{E}_c\) was distilled to only reflect outdated info like: \textit{"According to retrieved text, Kareem Abdul-Jabbar is the top scorer."}
\newline
The outputs are:
Conflict Flag ($\delta_c$): $1$.
Conflict Rationale ($r_c$): \textit{"parameter knowledge (\(\mathcal{E}_p\)) states LeBron James, while context-derived information (\(\mathcal{E}_c\)) states Kareem Abdul-Jabbar. These conflict."}

\item \textbf{CARE-RAG Generation (generates $\hat{a}$)}:
The LLM \(\mathcal{M}\), using prompt \(\Pi_{\text{synth}}\) (Appendix~\ref{app:prompt:reflective_synthesis}), receives \(q\), \(\mathcal{E}_p\), \(\mathcal{E}_c\), \(\delta_c=1\), and \(r_c\).
The \textbf{Final Answer ($\hat{a}$)} is, for example: \textit{"LeBron James is NBA's all-time leading scorer. While some historical records mention Kareem Abdul-Jabbar, LeBron James has surpassed this record, aligning with current information."}
\newline
The \textbf{Reasoning} would acknowledge the conflict identified by \(r_c\) and explain the prioritization of current parameter knowledge (\(\mathcal{E}_p\)) or the more recent parts of \(\mathcal{E}_c\), treating Kareem's record as historical.

\end{enumerate}
\section{Component Output Examples}
\label{app:component_examples} 
\vspace{-0.5\baselineskip}

This section provides additional, isolated examples of outputs from key components and stages of the Conflict-Aware and Reliable Evidence for RAG (CARE-RAG) framework. These examples illustrate the specific outputs for Context-aware evidence Generation (formerly Structured Reasoning), Conflict Detection, and CARE-RAG Generation. Examples for QA Repair (Appendix~\ref{app:qarepair}) and parameter-aware evidence Generation (\(\mathcal{E}_p\), detailed in Appendix~\ref{app:prompt:parameter_record}) are covered elsewhere or are straightforward.

\begin{tcolorbox}[
  colframe=red!75, 
  colback=red!5,
  boxrule=0.5pt,
  arc=2pt,
  width=\columnwidth,
  fontupper=\footnotesize,
  sharp corners=south,
  left=2mm, right=2mm, top=1mm, bottom=1mm,
  before skip=4pt, after skip=4pt,
  title=\textbf{E.1 \ding{72} Conflict Detection Output Example ($\delta_c, r_c$ from $\mathcal{M}_c$)}
]
\smallskip
\noindent \textbf{\textcolor{myorange}{Task}}: Evaluate whether the consolidated parameter-aware evidences ($\mathcal{E}_p$) contradict the refined Context-aware evidence ($\mathcal{E}_c$) for the given query. Output a conflict flag ($\delta_c \in \{0,1\}$) and a rationale ($r_c$). This uses prompt $\Pi_c$ (Appendix~\ref{app:prompt:conflict_detection}).

\vspace{0.5em}
\noindent \textbf{\textcolor{myorange}{Query ($q$)}}: \quad \textit{Who was "Suite: Judy Blue Eyes" written about?}

\vspace{0.5em}
\noindent \textbf{\textcolor{myorange}{Input: Consolidated parameter-aware evidences ($\mathcal{E}_p$) (simulated)}}:
\begin{itemize}[leftmargin=1.2em, itemsep=0.1em]
    \item \textit{Stephen Stills wrote it about Judy Collins, his former girlfriend.}
\end{itemize}

\vspace{0.5em}
\noindent \textbf{\textcolor{myorange}{Input: Refined Context-aware evidence ($\mathcal{E}_c$) (simulated)}}:
\begin{itemize}[leftmargin=1.2em, itemsep=0.1em]
    \item \textit{The song "Suite: Judy Blue Eyes" was written by Stephen Stills. It references Judy Collins and their relationship.}
\end{itemize}

\vspace{0.5em}
\noindent \textbf{\textcolor{mypurple}{LLM Reasoning for Conflict Detection (part of $r_c$)}}:
\begin{itemize}[leftmargin=1.2em, itemsep=0.1em]
    \item Both inputs identify Stephen Stills as the author and Judy Collins as the subject.
    \item The information provided is consistent and complementary, with no factual contradictions.
\end{itemize}

\vspace{0.5em}
\noindent \textbf{\textcolor{myred}{Conflict Detection Output}}:
\begin{itemize}[leftmargin=1.2em, itemsep=0.1em]
    \item \textbf{Conflict Flag ($\delta_c$)}: \quad \textbf{0}
    \item \textbf{Conflict Rationale ($r_c$)}: \quad \textit{No conflict detected. Both parameter knowledge and context-derived information consistently identify Stephen Stills as the author and Judy Collins as the subject of the song.}
\end{itemize}
\end{tcolorbox}

\vspace{1em}

\begin{tcolorbox}[
  colframe=blue!70!black, 
  colback=blue!5,
  boxrule=0.4pt,
  arc=2pt,
  width=\columnwidth,
  fontupper=\footnotesize,
  sharp corners=south,
  left=2mm, right=2mm, top=1mm, bottom=1mm,
  before skip=4pt, after skip=4pt,
  title=\textbf{E.2 \ding{44} Context-aware evidence Generation Example ($\mathcal{E}_c$ from $\Pi_{\text{ref}}$)}
]
\smallskip
\noindent \textbf{\textcolor{myorange}{Task}}: Distill the retrieved context ($C$) into a concise and salient Context-aware evidence ($\mathcal{E}_c$). This uses prompt $\Pi_{\text{ref}}$ (Appendix~\ref{app:prompt:retrieval_refinement}).

\vspace{0.5em}
\noindent \textbf{\textcolor{myorange}{Input Question ($q$)}}: \textit{Who was Henry Feilden?}

\vspace{0.5em}
\noindent \textbf{\textcolor{myorange}{Input Retrieved Context ($C$)}}:
\begin{itemize}[leftmargin=1.2em, itemsep=0.1em]
    \item \textit{"Colonel Henry Wemyss Feilden CB (1838–1921) was a British Army officer, naturalist and Arctic explorer."}
    \item \textit{"Feilden was the second son of the second Baronet Feilden..."}
    \item \textit{"Henry Master Feilden (1818–1875) was a Conservative politician..."}
    \item \textit{"Some sources incorrectly identify Henry Feilden as an actor."}
\end{itemize}

\vspace{0.5em}
\noindent \textbf{\textcolor{mypurple}{Generated Context-aware evidence ($\mathcal{E}_c$)}}:
\begin{itemize}[leftmargin=1.2em, itemsep=0.1em]
    \item \textbf{Key Factual Claims}:
        \begin{itemize}[leftmargin=1em]
            \item Two distinct individuals named Henry Feilden are mentioned:
                \begin{itemize}[leftmargin=1em]
                    \item Colonel Henry Wemyss Feilden (1838–1921): British Army officer, naturalist, Arctic explorer.
                    \item Henry Master Feilden (1818–1875): Conservative politician.
                \end{itemize}
            \item The claim that a Henry Feilden was an actor is explicitly refuted for at least one, and implicitly for both if the context is considered complete regarding their professions.
        \end{itemize}
    \item \textbf{Identified Ambiguities/Issues}: The query "Who was Henry Feilden?" is ambiguous without further specification, as the context provides two distinct individuals.
    \item \textbf{Context-Grounded Conclusion/Summary}: The retrieved context describes two notable individuals named Henry Feilden. To provide a specific answer to "Who was Henry Feilden?", clarification on which individual is being referred to would be necessary. The claim about Henry Feilden being an actor is noted as incorrect based on the context.
\end{itemize}
\end{tcolorbox}

\vspace{1em}

\begin{tcolorbox}[
  colframe=green!60!black, 
  colback=green!5,
  boxrule=0.4pt,
  arc=2pt,
  width=\columnwidth,
  fontupper=\footnotesize,
  sharp corners=south,
  left=2mm, right=2mm, top=1mm, bottom=1mm,
  before skip=4pt, after skip=4pt,
  title=\textbf{E.3 \ding{115} CARE-RAG Generation Output Example ($\hat{a}$ from $\Pi_{\text{synth}}$)}
]
\smallskip
\noindent \textbf{\textcolor{myorange}{Task}}: Generate the final reconciled answer ($\hat{a}$) by integrating all available information. This uses prompt $\Pi_{\text{synth}}$ (Appendix~\ref{app:prompt:reflective_synthesis}).

\vspace{0.5em}
\noindent \textbf{\textcolor{myorange}{Input Question ($q$)}}: \quad Who is the original artist of \textit{Sound of Silence}, released in 1964?

\vspace{0.5em}
\noindent \textbf{\textcolor{myorange}{Input: Consolidated parameter-aware evidences ($\mathcal{E}_p$) (simulated)}}: \textit{Simon \& Garfunkel}

\vspace{0.5em}
\noindent \textbf{\textcolor{myorange}{Input: Refined Context-aware evidence ($\mathcal{E}_c$) (simulated)}}: \textit{Based on the provided context, Simon \& Garfunkel are identified as the original artists of "Sound of Silence" released in 1964. No conflicting information found in context.}

\vspace{0.5em}
\noindent \textbf{\textcolor{myorange}{Input: Conflict Flag ($\delta_c$)}}: $0$
\noindent \textbf{\textcolor{myorange}{Input: Conflict Rationale ($r_c$)}}: \textit{No conflict detected between parameter knowledge and context-derived information.}

\vspace{0.5em}
\noindent \textbf{\textcolor{mypurple}{CARE-RAG Generation Output ($\hat{a}$)}}:
\begin{itemize}[leftmargin=1.2em, itemsep=0.1em]
    \item \textbf{Final Answer}: Simon \& Garfunkel were the original artists of the song "Sound of Silence," released in 1964.
    \item \textbf{Reasoning for Final Answer}: Both the consolidated parameter knowledge (\(\mathcal{E}_p\)) and the refined context-aware evidence (\(\mathcal{E}_c\)) consistently identify Simon \& Garfunkel. The conflict flag ($\delta_c=0$) confirms no discrepancy was found. There is no ambiguity regarding the 1964 release.
    \item \textbf{Ambiguity/Uncertainty Assessment}: None detected.
\end{itemize}
\end{tcolorbox}

\end{document}